\pdfoutput=1
\documentclass[11pt]{article}

\usepackage[preprint]{coling}

\usepackage{times}
\usepackage{latexsym}
\usepackage{xcolor}

\usepackage[T1]{fontenc}
\usepackage[utf8]{inputenc}

\usepackage{microtype}

\usepackage{inconsolata}

\usepackage{graphicx}

\usepackage{xcolor}
\usepackage{subcaption}
\usepackage{multicol}
\usepackage{url}

\title{Assessing Open-Source Large Language Models on Argumentation Mining Subtasks}

\author{
  \textbf{Mohammad Yeghaneh Abkenar $^*$\textsuperscript{1,3}}\\
  \texttt{yeghanehabkenar@uni-potsdam.de} \And
  \textbf{Weixing Wang$^*$\textsuperscript{2,3}} \\
  \texttt{weixing.wang@hpi.de} 
  \AND
  \textbf{Hendrik Graupner\textsuperscript{1,2,3}} \\
  \texttt{hendrik.graupner@hpi.de} \And
  \textbf{Manfred Stede\textsuperscript{3}} \\
  \texttt{stede@uni-potsdam.de}
\\
\AND
  \textsuperscript{1}Bundesdruckerei Gruppe GmbH Berlin,
  \textsuperscript{2}Hasso Plattner Institute,
  \textsuperscript{3}University of Potsdam
\\
}
\begin{document}
\maketitle

\begin{abstract}
We explore the capability of four open-source large language models (LLMs) in argumentation mining (AM). We conduct experiments on three different corpora; persuasive essays (PE), argumentative microtexts (AMT) Part 1 and Part 2, based on two argumentation mining sub-tasks: (i) argumentative discourse units classifications (ADUC), and (ii) argumentative relation classification (ARC). This work aims to assess the argumentation capability of open-source LLMs, including Mistral 7B, Mixtral 8x7B, LlamA2 7B and LlamA3 8B in both, zero-shot and few-shot scenarios. Our analysis contributes to
further assessing computational argumentation with open-source LLMs in future research efforts\footnote{Code and data available on \url{https://anonymous.4open.science/r/openarg-41C0}}.

\end{abstract}
\section{Introduction}

Over the past few years, advancements in the broader field of natural language processing (NLP), such as pre-trained transformer-based models \cite{devlin2018bert}, coupled with the increasing availability of diverse data, have significantly enhanced the potential for nearly every area of NLP, including argumentation mining (AM) \cite{stede-schneider-2018, lawrence-reed-2020}. AM, and specifically the problem of finding argumentation structures in text, has received much attention in the past decade. The objective of AM is to detect argumentation within text or dialogue, to create detailed representations of claims and their supporting or attacking arguments, and to analyze the reasoning patterns that validate the argumentation. Beyond academic interest, AM attracts significant attention for its diverse applications, as demonstrated by projects like IBM Debater \cite{bar2021project}, decision assistance \cite{liebeck2016airport}, product reviews \cite{passon2018predicting} and writing support \cite{wachsmuth2016using}.

\section{Background and Related work} 

\subsection{Argumentation Mining}
Unlike many NLP problems, argumentation mining (AM) is not a single, straightforward task but rather a collection of interrelated subtasks. AM enhances sentiment analysis by delving deeper into the reasoning behind opinions. While sentiment analysis identifies "what people think about entity X," AM explores "why people think Y about X."
One sub-task we address is argumentative discourse unit classification (ADUC), which identifies the type of argumentative discourse units, as defined by
\citet[p.~14]{hidey2017analyzing} as as follows:
\begin{itemize}
    \item{\textit{Claim} (Conclusion): A statement in the text that articulates a perspective on a particular issue. It can include predictions, interpretations, evaluations, and expressions of agreement or disagreement with others' assertions.}
    \item{\textit{Premise} (Evidence): A statement presented to strengthen a claim, designed to persuade the audience of its validity. Although premises may express opinions, their main function is to support or refute an existing proposition rather than introduce a new perspective.}
\end{itemize}

 We also cover argumentative relation classification (ARC) to
identify relations among argumentative discourse units (ADUs) which is defined by  \citet[p.~491]{ali2022constructing}  as follows:

\begin{itemize}
    \item{\textit{Support} (For): 
    The Support relation occurs when a premise enhances or reinforces a claim. This can happen in various ways: If the claim is a proposition (such as a fact, opinion, or belief), the premise strengthens the claim’s likelihood or truth. If the claim is an action, the premise provides justification or makes the action more acceptable. If the claim is an event, the premise increases the probability that the event occurred.}
    
    \item{\textit{Attack} (Against): 
    The Attack relation occurs when a premise undermines or contradicts a claim. This can manifest in several ways: If the claim is a proposition, the premise weakens the claim’s likelihood or truth. If the claim is an action, the premise denies or challenges the justification for the action. If the claim is an event, the premise reduces the probability that the event occurred.}
\end{itemize}

\subsection{Using LLMs for AM}

Recently, we saw huge breakthroughs in language modeling. Large Language Models such as GPT4 \citep{achiam2023gpt}, Llama3 \citep{dubey2024llama}, and Mistral \citep{jiang2023mistral} have demonstrated strong capabilities in solving various NLP tasks. 
%LLMs are capable of capturing the nuances, context, and semantics of the human language, allowing them to perform tasks such as text generation \citep{zhao2023survey}, summarization \citep{jin2024comprehensive, chang2023booookscore, zhang2024benchmarking}, translation \citep{wu2024adapting, xu2024contrastive, li2024eliciting}, question answering \citep{li2024flexkbqa, wei2022chain}, and more. 
As a result, there is an increasing interest in applying LLMs for computational argumentation tasks. For example, \citet{de2023wish} evaluated the ability of two LLMs to perform argumentative reasoning. Their experiments involved argumentation mining and argument pair extraction, assessing the LLMs’ capability to recognize arguments under progressively more abstract input and output representations. However, their research is limited to the two closed-source language models GPT3 and GPT4. \citet{chen2023exploring} conducted a comprehensive analysis of LLMs on diverse computational argumentation tasks, their goal was to evaluate LLMs including ChatGPT, Flan models, and LLaMA2 models in both zero-shot and few-shot settings. However, their studies did not address the argumentative relation classification sub-task and they did not use some state-of-the-art models such as LLaMA3 and the Mistral family which according to \citet{sinha2024evaluating} are also promising in various reasoning tasks.

To overcome the above limitations, we explore two key sub-tasks of argumentation: argumentation discourse unit classification and argument relation classification, using four open-source LLMs across three well-known argument mining corpora. We believe that argumentation mining sub-tasks are fundamentally different from argument pair extraction and argument generation. As such, argumentation mining sub-tasks need to be explored differently using various LLMs on the most well-known and important corpora with a similar structure.

\section{Corpora and Task Definition}

%One approach to assessing the reasoning capabilities of LLMs is to evaluate concretely their performance on tasks that necessitate reasoning. We have chosen this approach, in order to measure the ability of different large language models in reasoning.
In this paper, we conduct experiments on two central sub-tasks of argumentation mining using three well-known datasets which will be introduced in the next sub-sections.

\subsection{Corpora}

\begin{table}[h!]
\centering
\resizebox{\columnwidth}{!}{%
\begin{tabular}{lccc|ccc}
  \hline
  \textbf{Dataset} & \multicolumn{3}{c|}{\textbf{ADUC}} & \multicolumn{3}{c}{\textbf{ARC}}  \\
  \hline
  & \textbf{Total} & \textbf{Claim} & \textbf{Premise} & \textbf{Total} & \textbf{Support} & \textbf{Attack} \\
  \hline
  AMT1 & 576 & 112 & 464 & 455 & 284 & 171 \\
  AMT2 & 932 & 171 & 761 & 738 & 524 & 214 \\ 
  PE & 6089 & 2257 & 3832 & 3821 & 3603 & 218 \\
  \hline
\end{tabular}%
}
\caption{Summary of sample number and label distributions of the three corpora.}
\label{tab:statistic}

\end{table}

\paragraph{Argumentative Microtexts (Part 1).}

The AMT1 corpus, created by \cite{peldszus2015annotated}, includes 112 short texts (each about 3–5 sentences long) and 576 argumentative discourse units. They were originally written in German and have been professionally translated to English,  as well as 
to Italian \cite{namor2019mining}, Russian \cite{fishcheva2019cross} and recently to Persian  \cite{abkenar2024neural} preserving the segmentation and if possible the usage of discourse markers and annotated with complete argumentation tree structures. 
\paragraph{Argumentative Microtexts (Part 2).}

The second part of AMT, created by \cite{skeppstedt2018more} using crowd-sourcing, includes 171 short texts with 932 argumentative discourse units 
in English which is annotated consistent with the approach utilized in the original corpus. One of the differences in this corpus is the existence of an implicit claim which is marked in the XML file. 

\paragraph{Persuasive Essays}
 The PE corpus comprises 402 argumentative essays (totaling 2235 paragraphs) written by English learners in response to specific prompts. \citet{stab2017parsing} collected these essays from a website and annotated them with argumentation graphs. The essays begin with a question and include a major claim supported by evidence, which may have a substructure. Some sentences are non-argumentative, providing only background or minor elaborations. Each essay has a major claim, typically found at the end, supported by claims within the paragraphs. For consistency with other corpora, we treat "major claim" and "claim" as equivalent and classify argument components (ACs) at the paragraph level.

\subsection{Tasks}
   
\paragraph{Argumentative Discourse Units Classification (ADUC)}
Argumentative discourse units (ADUs) are minimal units of analysis, i.e., the smallest elements in a text that contribute to argumentative structure.  In this paper we define ADUC as the classification of these units as either "premise" or "claim"; we do not address the distinction between ADUs and non-argumentative material.

\paragraph{Argumentative Relation Classification (ARC)} 

The goal of argumentative relation identification is to determine whether each pair of ADUs is argumentatively related or not \cite{rocha2018cross}.  We assume that the task of segmenting the text into ADUs has already been completed. Given a pair of ADUs, the objective is to classify the relation between them as either "support" or "attack."

\section{Methods}

\begin{figure}[!ht]
\centering
\includegraphics[width = 1.1\columnwidth]{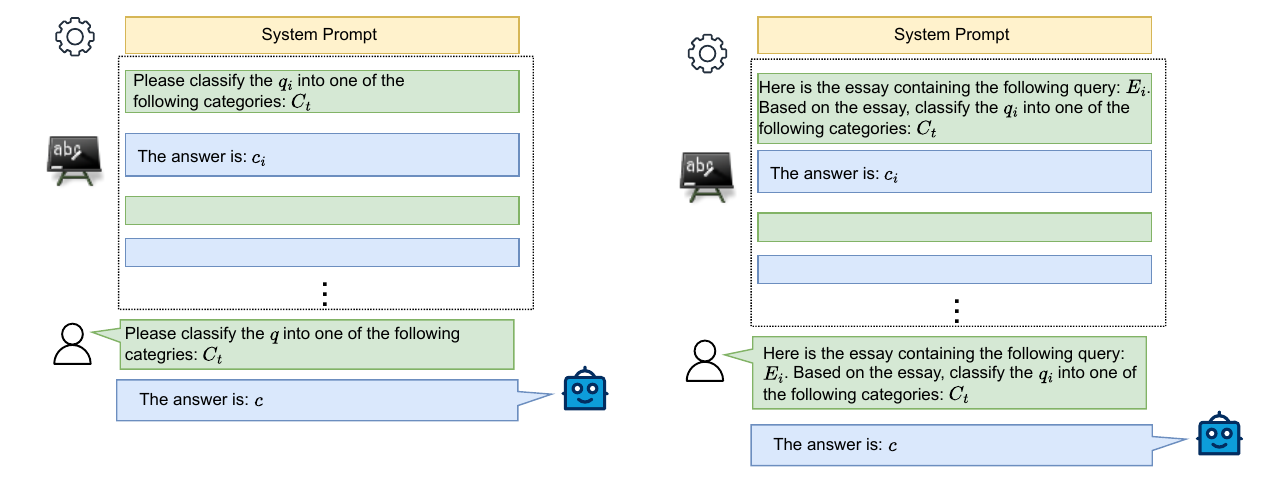} 
\caption{An overview of the prompting methods. Left: Vanilla Prompting. Right: Context-Aware Prompting}
\label{fig:prompt_design}
\end{figure}.

\subsection{Vanilla Prompting}
This approach involves asking the model to classify each ADU independently, without considering the whole context. As you see on the left side of Figure~\ref{fig:prompt_design}, we ask the model: "Please classify the following ADU $q_{i}$ into one of the categories $C_{i}$." This is the same for the ARC, but we ask the same question on pairs of ADUs.

\subsection{Context-Aware Prompting}

This approach asks the model to classify each ADU based on its context in the text. As shown on the right side of Figure~\ref{fig:prompt_design}, we prompt the model with: 

"Here is the essay containing the following query $E_{i}$.
Based on the essay, classify $q_{i}$ into one of the categories $C_{i}$."
Unlike the standard method, where each ADU is classified independently, this context-aware prompting requires the model to consider the surrounding context of the essay or microtext for each ADU. For ARC, we ask the model to classify pairs of ADUs, still taking into account the context provided. See appendix \ref{sec:prompt} for more.

\section{Experiments and Results}
\subsection{Zero-Shot Performance}

\begin{table}[h!]
\centering
\resizebox{\columnwidth}{!}{%
\begin{tabular}{lcc|cc|cc}
  \hline
  \textbf{Model} & \multicolumn{2}{c|}{\textbf{AMT1}} & \multicolumn{2}{c|}{\textbf{AMT2}} & \multicolumn{2}{c}{\textbf{PE}} \\
  \hline
  & \textbf{ADUC} & \textbf{ARC} & \textbf{ADUC} & \textbf{ARC} & \textbf{ADUC} & \textbf{ARC} \\
  \hline
  Mistral (Vanilla) & \textbf{0.746} & 0.491 & 0.656 & 0.578 & 0.623 & 0.724 \\
  Mistral (Context) & 0.463 & \textbf{0.651} & 0.456 & 0.693 & 0.475 & 0.792 \\ \hline
  Mixtral (Vanilla) & 0.728 & 0.556 & 0.566 & 0.638 & 0.551 & 0.784 \\
  Mixtral (Context) & 0.585 & 0.604 & 0.598 & \textbf{0.734} & 0.499 & 0.887 \\ \hline
  Llama2 (Vanilla) & 0.489 & 0.216 & 0.465 & 0.291 & 0.571 & 0.350 \\
  Llama2 (Context) & 0.574 & 0.017 & 0.577 & 0.222 & \textbf{0.696} & 0.632 \\ \hline
  Llama3 (Vanilla) & 0.718 & 0.222 & 0.514 & 0.703 & 0.634 & 0.883 \\
  Llama3 (Context) & 0.657 & 0.302 & \textbf{0.671} & 0.719 & 0.588 & \textbf{0.931} \\
  \hline
\end{tabular}%
}
\caption{Performance of different models across AMT1, AMT2 and PE corpora on ADUC, and ARC tasks.}
\label{tab:zeroshot-performance}

\end{table}

We test our two prompting methods with 4~advanced LLMs, namely  \textbf{Llama 2-7B} \citep{touvron2023llama}, \textbf{Llama 3-8B} \citep{dubey2024llama3herdmodels}, \textbf{Mistral-7B} \citep{jiang2023mistral}, and \textbf{Mixtral-8x7B} \citep{jiang2024mixtralexperts}. For comparison, we report the micro F\textsubscript{1} score, because the datasets are all imbalanced.

Table~\ref{tab:zeroshot-performance} presents the results of zero-shot learning. When comparing ADUC and ARC tasks, we find that context prompts generally improve ARC performance across most models and datasets, suggesting that context aids in better understanding relationships between sentences. However, the ADUC task appears more sensitive to the introduction of context, with some models experiencing a performance drop. Llama3 stands out for maintaining strong performance across both tasks and all datasets when using context prompts. This suggests that Llama3 is more adaptable to varying prompting methods and datasets in AM tasks.

\subsection{Performance and Number of Demonstrations}
% Additional text or analysis can follow here.
\begin{figure}[htb!]
    \centering
    \begin{minipage}{0.72\columnwidth}
        \centering
        \includegraphics[width=\textwidth]{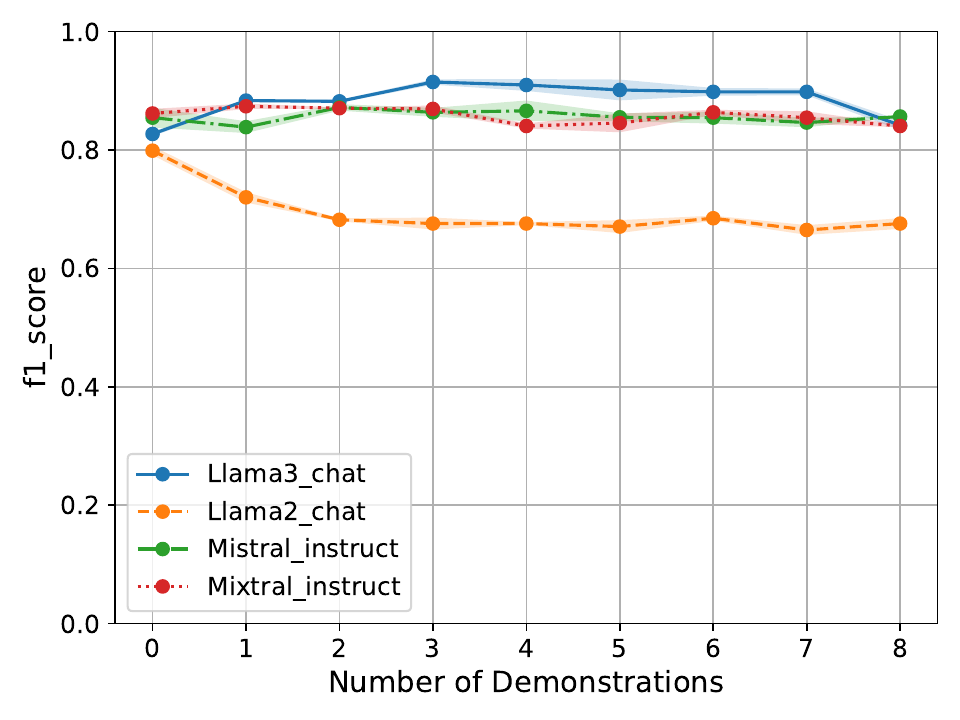}
        \caption*{(a) AMT1 vanilla prompting}
    \end{minipage}
    \hfill
    \begin{minipage}{0.72\columnwidth}
        \centering
        \includegraphics[width=\textwidth]{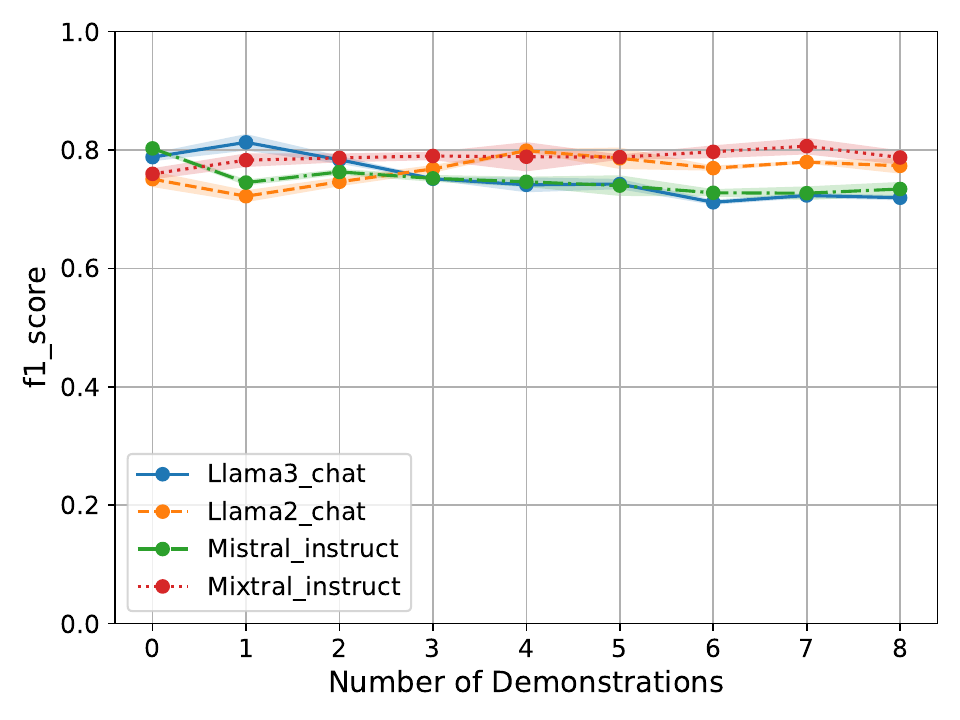}
        \caption*{(b) AMT1 context-aware prompting}
    \end{minipage}
    \caption{Few-shot performance on ADUC task with the AMT1 dataset.}
    \label{fig:adu_classify}
\end{figure}
Figure~\ref{fig:adu_classify} illustrates the performance on the ADUC task across the datasets under different numbers of demonstrations. For the ADUC task, context-aware prompting can bring all models to a similar level. I.e., weaker models like Llama2 are enhanced while the stronger models are degraded. For example, considering three-shot learning on the AMT1 dataset, Llama3 can achieve 86\% accuracy using vanilla prompting (a), but the accuracy drops to 61\% with context-aware prompting (b). In comparison, Llama2 improves from 48\% (a) to 63\% (d) by applying context-aware prompting. Moreover, we find that the application of context-aware prompting significantly reduces the performance disparity between the AMT1 and AMT2 datasets. For the complete comparison please refer to appendix Section~\ref{sec:results}. This suggests that providing additional contextual information helps the models to handle variations between these datasets more effectively, resulting in a more uniform performance across different versions of the AM tasks.

\begin{figure}[ht!]
    \centering
    \begin{minipage}{0.72\columnwidth}
        \centering
        \includegraphics[width=\textwidth]{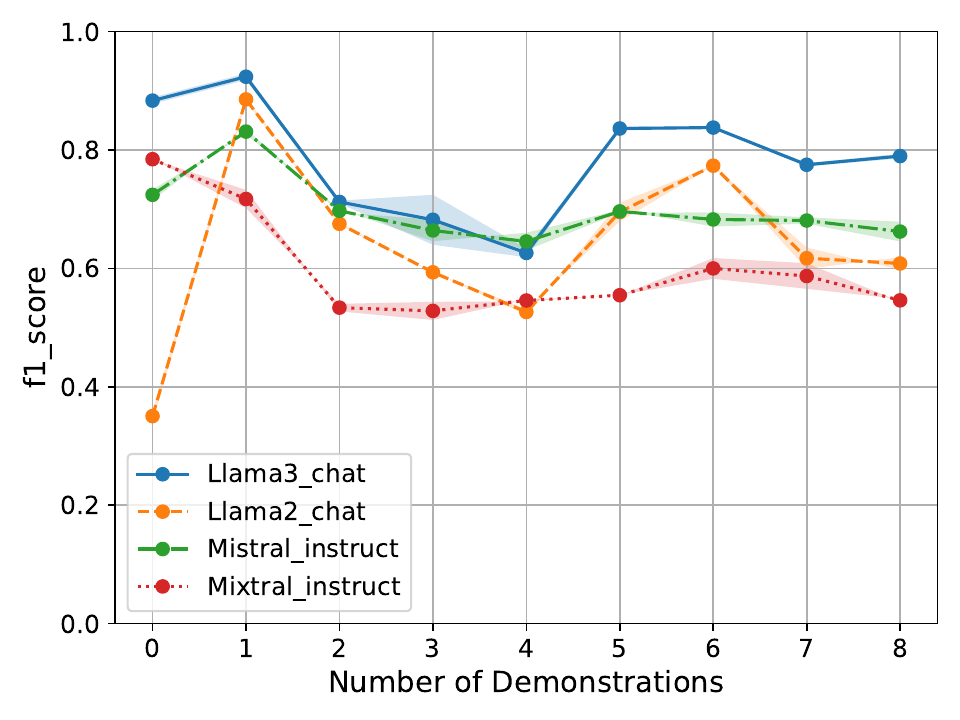}
        \caption*{(a) PE vanilla prompting}
    \end{minipage}
    \hfill
    \begin{minipage}{0.72\columnwidth}
        \centering
        \includegraphics[width=\textwidth]{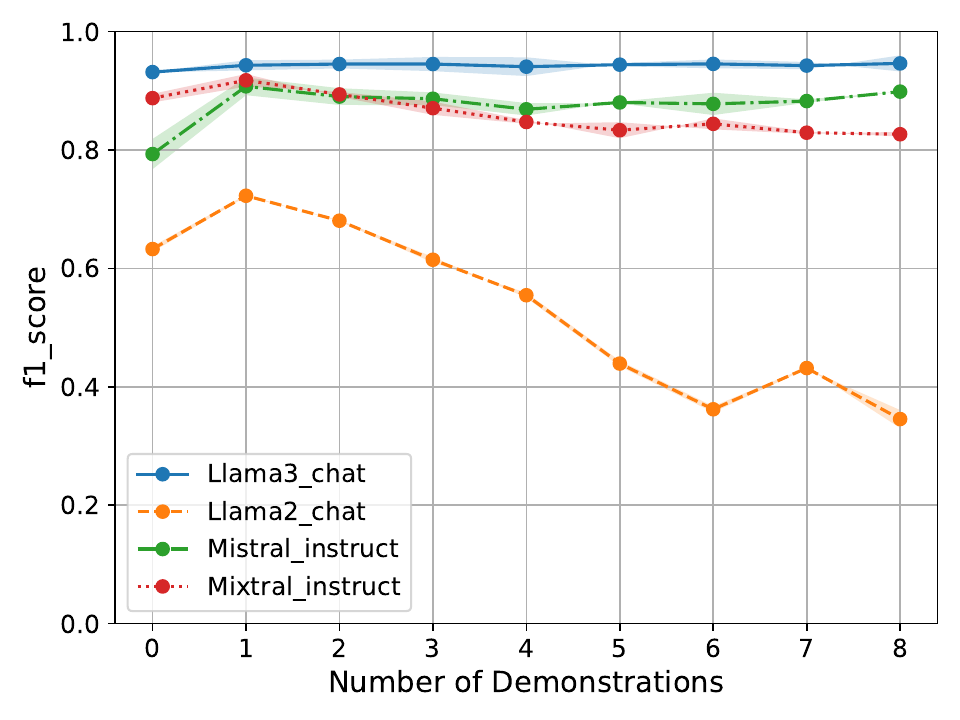}
        \caption*{(b) PE context-aware prompting}
    \end{minipage}

    \caption{Few-shot performance on ARC task with the PE dataset. }
    \label{fig:ar_classify}
\end{figure}

For the ARC task, we see slightly different patterns of model performance in Figure~\ref{fig:ar_classify}. However, we still observe that context-aware prompting serves as an effective stabilizer for model performance. Comparing (a) and (b), we find when models are prompted with additional contextual information, they exhibit reduced fluctuations in their performance regarding different numbers of demonstrations, suggesting that this approach helps mitigate the impact of noise brought by additional demonstrations. In contrast, vanilla prompting, which lacks this additional context, often results in more erratic performance across different numbers of demonstrations, likely because the models are more susceptible to the inherent variability and difficulty of the tasks. This fluctuation in vanilla prompting can be attributed to the models' struggle to consistently grasp the underlying patterns in the data without sufficient context, leading to inconsistent F\textsubscript{1} scores. By providing context-aware prompting, the models are better equipped to understand and process the tasks at hand, resulting in more stable and reliable outputs.

%\begin{figure}[!ht]
%\centering
%\includegraphics[width=\columnwidth]{figures/pe_st_f1.pdf} 
%\caption{ST classification for PE dataset}
%\label{fig:pe_st}
%\end{figure}

\section{Discussion and Conclusion}

We assessed the reasoning abilities of four LLMs. Our evaluation focused on two key tasks in argumentation mining: ADUC and ARC.  The LLMs performed well in ADUC and particularly excelled in ARC, surpassing or closely matching the state-of-the-art results reported in \cite{abkenar2021neural} and \cite{chernodub2019targer} for AMT1  and PE, based on the F\textsubscript{1} micro score. However, statistical analysis of the LLMs' predictions shows that LLMs generally perform worse on AMT2 than AMT1, which may be because of the lower quality of AMT2. We also revealed that demonstrations serve as stabilizers rather than enhancers for both AM tasks.

\section{Limitation and Future Work}

We conducted our study on two central sub-tasks of argumentation mining (AM). However, other sub-tasks, such as the identification of argument components and the evaluation of argument quality, need to be addressed. Additionally, our results focus exclusively on English argumentative corpora. We recommend that future research explore other languages, particularly those with lower resources.

%\section*{Acknowledgments}

%We would like to express our appreciation to the Bundesdruckerei-Gruppe and the Hasso Plattner Institute for providing us with the opportunity to freely work on our research topics. Thank you for fostering an environment that encourages innovation and academic growth.

\bibliography{custom}

\begin{thebibliography}{27}
\providecommand{\natexlab}[1]{#1}

\bibitem[{Abkenar and Stede(2024)}]{abkenar2024neural}
Mohammad~Yeghaneh Abkenar and Manfred Stede. 2024.
\newblock Neural mining of persian short argumentative texts.
\newblock In \emph{Proceedings of the 2nd Workshop on Resources and Technologies for Indigenous, Endangered and Lesser-resourced Languages in Eurasia (EURALI)@ LREC-COLING 2024}, pages 30--35.

\bibitem[{Abkenar et~al.(2021)Abkenar, Stede, and Oepen}]{abkenar2021neural}
Mohammad~Yeghaneh Abkenar, Manfred Stede, and Stephan Oepen. 2021.
\newblock Neural argumentation mining on essays and microtexts with contextualized word embeddings.

\bibitem[{Achiam et~al.(2023)Achiam, Adler, Agarwal, Ahmad, Akkaya, Aleman, Almeida, Altenschmidt, Altman, Anadkat et~al.}]{achiam2023gpt}
Josh Achiam, Steven Adler, Sandhini Agarwal, Lama Ahmad, Ilge Akkaya, Florencia~Leoni Aleman, Diogo Almeida, Janko Altenschmidt, Sam Altman, Shyamal Anadkat, et~al. 2023.
\newblock Gpt-4 technical report.
\newblock \emph{arXiv preprint arXiv:2303.08774}.

\bibitem[{Ali et~al.(2022)Ali, Pawar, Palshikar, and Singh}]{ali2022constructing}
Basit Ali, Sachin Pawar, Girish Palshikar, and Rituraj Singh. 2022.
\newblock Constructing a dataset of support and attack relations in legal arguments in court judgements using linguistic rules.
\newblock In \emph{Proceedings of the Thirteenth Language Resources and Evaluation Conference}, pages 491--500.

\bibitem[{Bar-Haim et~al.(2021)Bar-Haim, Kantor, Venezian, Katz, and Slonim}]{bar2021project}
Roy Bar-Haim, Yoav Kantor, Elad Venezian, Yoav Katz, and Noam Slonim. 2021.
\newblock Project debater apis: Decomposing the ai grand challenge.
\newblock \emph{arXiv preprint arXiv:2110.01029}.

\bibitem[{Chen et~al.(2023)Chen, Cheng, Tuan, and Bing}]{chen2023exploring}
Guizhen Chen, Liying Cheng, Luu~Anh Tuan, and Lidong Bing. 2023.
\newblock Exploring the potential of large language models in computational argumentation.
\newblock \emph{arXiv preprint arXiv:2311.09022}.

\bibitem[{Chernodub et~al.(2019)Chernodub, Oliynyk, Heidenreich, Bondarenko, Hagen, Biemann, and Panchenko}]{chernodub2019targer}
Artem Chernodub, Oleksiy Oliynyk, Philipp Heidenreich, Alexander Bondarenko, Matthias Hagen, Chris Biemann, and Alexander Panchenko. 2019.
\newblock Targer: Neural argument mining at your fingertips.
\newblock In \emph{Proceedings of the 57th Annual Meeting of the Association for Computational Linguistics: System Demonstrations}, pages 195--200.

\bibitem[{de~Wynter and Yuan(2023)}]{de2023wish}
Adrian de~Wynter and Tommy Yuan. 2023.
\newblock I wish to have an argument: Argumentative reasoning in large language models.
\newblock \emph{arXiv preprint arXiv:2309.16938}.

\bibitem[{Devlin(2018)}]{devlin2018bert}
Jacob Devlin. 2018.
\newblock Bert: Pre-training of deep bidirectional transformers for language understanding.
\newblock \emph{arXiv preprint arXiv:1810.04805}.

\bibitem[{Dubey et~al.(2024{\natexlab{a}})Dubey, Jauhri, Pandey, Kadian, Al-Dahle, Letman, Mathur, Schelten, Yang, and Angela~Fan}]{dubey2024llama3herdmodels}
Abhimanyu Dubey, Abhinav Jauhri, Abhinav Pandey, Abhishek Kadian, Ahmad Al-Dahle, Aiesha Letman, Akhil Mathur, Alan Schelten, Amy Yang, and et~al. Angela~Fan. 2024{\natexlab{a}}.
\newblock \href {https://arxiv.org/abs/2407.21783} {The llama 3 herd of models}.
\newblock \emph{Preprint}, arXiv:2407.21783.

\bibitem[{Dubey et~al.(2024{\natexlab{b}})Dubey, Jauhri, Pandey, Kadian, Al-Dahle, Letman, Mathur, Schelten, Yang, Fan et~al.}]{dubey2024llama}
Abhimanyu Dubey, Abhinav Jauhri, Abhinav Pandey, Abhishek Kadian, Ahmad Al-Dahle, Aiesha Letman, Akhil Mathur, Alan Schelten, Amy Yang, Angela Fan, et~al. 2024{\natexlab{b}}.
\newblock The llama 3 herd of models.
\newblock \emph{arXiv preprint arXiv:2407.21783}.

\bibitem[{Fishcheva and Kotelnikov(2019)}]{fishcheva2019cross}
Irina Fishcheva and Evgeny Kotelnikov. 2019.
\newblock Cross-lingual argumentation mining for russian texts.
\newblock In \emph{International Conference on Analysis of Images, Social Networks and Texts}, pages 134--144. Springer.

\bibitem[{Hidey et~al.(2017)Hidey, Musi, Hwang, Muresan, and McKeown}]{hidey2017analyzing}
Christopher Hidey, Elena Musi, Alyssa Hwang, Smaranda Muresan, and Kathy McKeown. 2017.
\newblock Analyzing the semantic types of claims and premises in an online persuasive forum.
\newblock In \emph{Proceedings of the 4th Workshop on Argument Mining}. Columbia Univ., New York, NY (United States).

\bibitem[{Jiang et~al.(2023)Jiang, Sablayrolles, Mensch, Bamford, Chaplot, Casas, Bressand, Lengyel, Lample, Saulnier et~al.}]{jiang2023mistral}
Albert~Q Jiang, Alexandre Sablayrolles, Arthur Mensch, Chris Bamford, Devendra~Singh Chaplot, Diego de~las Casas, Florian Bressand, Gianna Lengyel, Guillaume Lample, Lucile Saulnier, et~al. 2023.
\newblock Mistral 7b.
\newblock \emph{arXiv preprint arXiv:2310.06825}.

\bibitem[{Jiang et~al.(2024)Jiang, Sablayrolles, Roux, Mensch, Savary, Bamford, Chaplot, de~las Casas, Hanna, Bressand, Lengyel, Bour, Lample, Lavaud, Saulnier, Lachaux, Stock, Subramanian, Yang, Antoniak, Scao, Gervet, Lavril, Wang, Lacroix, and Sayed}]{jiang2024mixtralexperts}
Albert~Q. Jiang, Alexandre Sablayrolles, Antoine Roux, Arthur Mensch, Blanche Savary, Chris Bamford, Devendra~Singh Chaplot, Diego de~las Casas, Emma~Bou Hanna, Florian Bressand, Gianna Lengyel, Guillaume Bour, Guillaume Lample, Lélio~Renard Lavaud, Lucile Saulnier, Marie-Anne Lachaux, Pierre Stock, Sandeep Subramanian, Sophia Yang, Szymon Antoniak, Teven~Le Scao, Théophile Gervet, Thibaut Lavril, Thomas Wang, Timothée Lacroix, and William~El Sayed. 2024.
\newblock \href {https://arxiv.org/abs/2401.04088} {Mixtral of experts}.
\newblock \emph{Preprint}, arXiv:2401.04088.

\bibitem[{Lawrence and Reed(2020)}]{lawrence-reed-2020}
John Lawrence and Chris Reed. 2020.
\newblock {Argument Mining: A Survey}.
\newblock \emph{Computational Linguistics}, 45(4):765--818.

\bibitem[{Liebeck et~al.(2016)Liebeck, Esau, and Conrad}]{liebeck2016airport}
Matthias Liebeck, Katharina Esau, and Stefan Conrad. 2016.
\newblock What to do with an airport? mining arguments in the german online participation project tempelhofer feld.
\newblock In \emph{Proceedings of the Third Workshop on Argument Mining (ArgMining2016)}, pages 144--153.

\bibitem[{Namor and Stede(2019)}]{namor2019mining}
Ivan Namor and Manfred Stede. 2019.
\newblock Mining italian short argumentative texts.
\newblock In \emph{Proceedings of the 5th Workshop on Argument Mining}.

\bibitem[{Passon et~al.(2018)Passon, Lippi, Serra, and Tasso}]{passon2018predicting}
Marco Passon, Marco Lippi, Giuseppe Serra, and Carlo Tasso. 2018.
\newblock Predicting the usefulness of amazon reviews using off-the-shelf argumentation mining.
\newblock \emph{arXiv preprint arXiv:1809.08145}.

\bibitem[{Peldszus and Stede(2015)}]{peldszus2015annotated}
Andreas Peldszus and Manfred Stede. 2015.
\newblock An annotated corpus of argumentative microtexts.
\newblock In \emph{Argumentation and Reasoned Action: Proceedings of the 1st European Conference on Argumentation, Lisbon}, volume~2, pages 801--815.

\bibitem[{Rocha et~al.(2018)Rocha, Stab, Cardoso, and Gurevych}]{rocha2018cross}
Gil Rocha, Christian Stab, Henrique~Lopes Cardoso, and Iryna Gurevych. 2018.
\newblock Cross-lingual argumentative relation identification: from english to portuguese.
\newblock In \emph{Proceedings of the 5th Workshop on Argument Mining, 2018 Conference on Empirical Methods in Natural Language Processing (EMNLP 2018)}.

\bibitem[{Sinha et~al.(2024)Sinha, Jain, and Chadha}]{sinha2024evaluating}
Neelabh Sinha, Vinija Jain, and Aman Chadha. 2024.
\newblock Evaluating open language models across task types, application domains, and reasoning types: An in-depth experimental analysis.
\newblock \emph{arXiv preprint arXiv:2406.11402}.

\bibitem[{Skeppstedt et~al.(2018)Skeppstedt, Peldszus, and Stede}]{skeppstedt2018more}
Maria Skeppstedt, Andreas Peldszus, and Manfred Stede. 2018.
\newblock More or less controlled elicitation of argumentative text: Enlarging a microtext corpus via crowdsourcing.
\newblock In \emph{Proceedings of the 5th Workshop on Argument Mining}, pages 155--163.

\bibitem[{Stab and Gurevych(2017)}]{stab2017parsing}
Christian Stab and Iryna Gurevych. 2017.
\newblock Parsing argumentation structures in persuasive essays.
\newblock \emph{Computational Linguistics}, 43(3):619--659.

\bibitem[{Stede and Schneider(2018)}]{stede-schneider-2018}
Manfred Stede and Jodi Schneider. 2018.
\newblock \emph{{Argumentation Mining}}, volume~40 of \emph{{Synthesis Lectures in Human Language Technology}}.
\newblock Morgan \& Claypool.

\bibitem[{Touvron et~al.(2023)Touvron, Martin, Stone, Albert, Almahairi, Babaei, Bashlykov, Batra, Bhargava, Bhosale et~al.}]{touvron2023llama}
Hugo Touvron, Louis Martin, Kevin Stone, Peter Albert, Amjad Almahairi, Yasmine Babaei, Nikolay Bashlykov, Soumya Batra, Prajjwal Bhargava, Shruti Bhosale, et~al. 2023.
\newblock Llama 2: Open foundation and fine-tuned chat models.
\newblock \emph{arXiv preprint arXiv:2307.09288}.

\bibitem[{Wachsmuth et~al.(2016)Wachsmuth, Al~Khatib, and Stein}]{wachsmuth2016using}
Henning Wachsmuth, Khalid Al~Khatib, and Benno Stein. 2016.
\newblock Using argument mining to assess the argumentation quality of essays.
\newblock In \emph{Proceedings of COLING 2016, the 26th international conference on Computational Linguistics: Technical papers}, pages 1680--1691.

\end{thebibliography}

\appendix

\section{Additional Results}
\label{sec:results}
\begin{figure*}[bht!]
    \centering
    \begin{minipage}{0.32\textwidth}
        \centering
        \includegraphics[width=\textwidth]{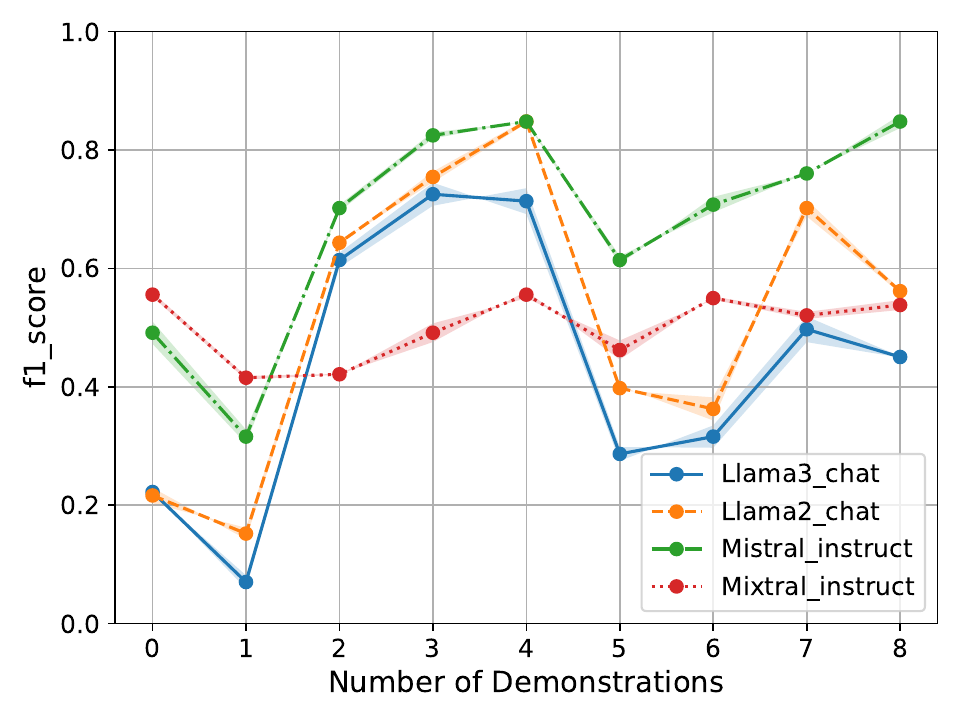}
        \caption*{(a) AMT1 vanilla prompting}
    \end{minipage}
    \hfill
    \begin{minipage}{0.32\textwidth}
        \centering
        \includegraphics[width=\textwidth]{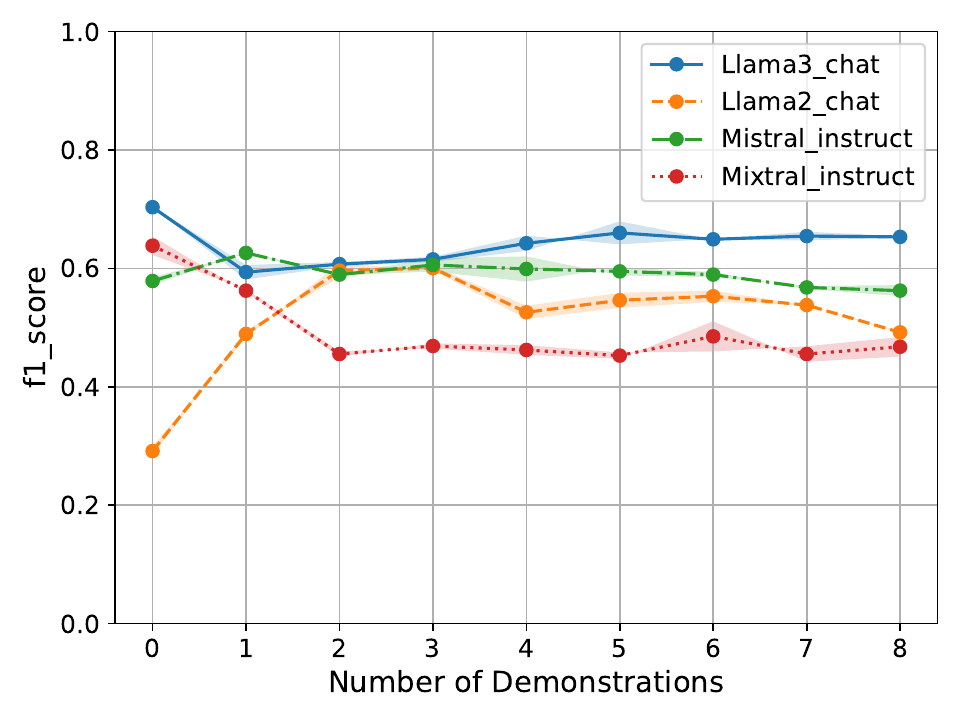}
        \caption*{(b) AMT2 vanilla prompting}
    \end{minipage}
    \hfill
    \begin{minipage}{0.32\textwidth}
        \centering
        \includegraphics[width=\textwidth]{new_figures/pe_sa_f1.pdf}
        \caption*{(c) PE vanilla prompting}
    \end{minipage}
    
    \vspace{10pt} % Adjust space between rows
    \begin{minipage}{0.32\textwidth}
        \centering
        \includegraphics[width=\textwidth]{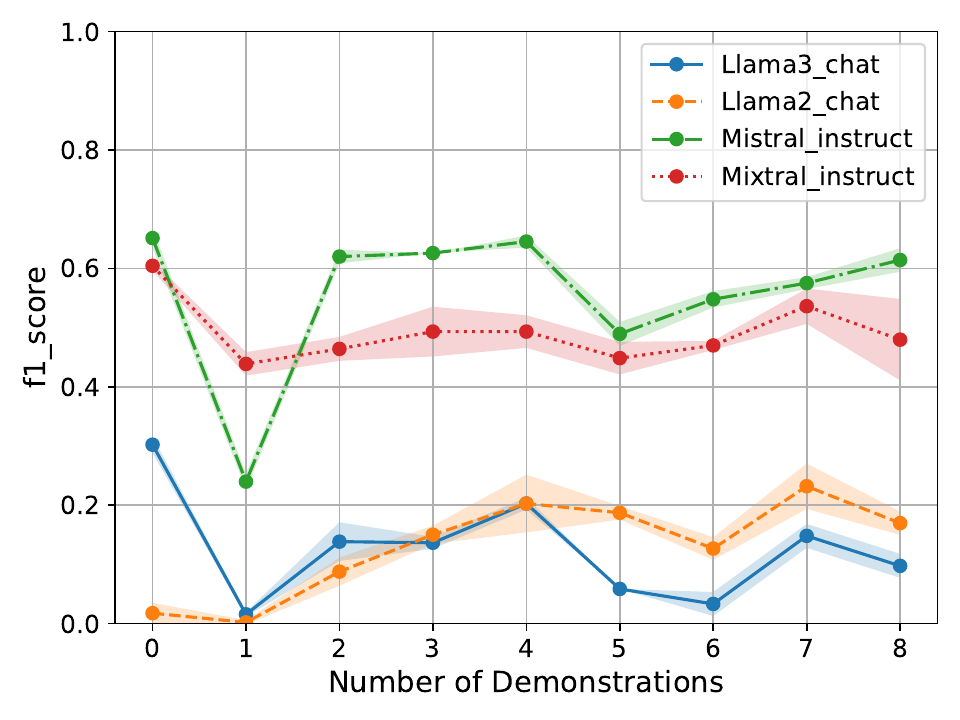}
        \caption*{(d) AMT1 context-aware prompting}
    \end{minipage}
    \hfill
    \begin{minipage}{0.32\textwidth}
        \centering
        \includegraphics[width=\textwidth]{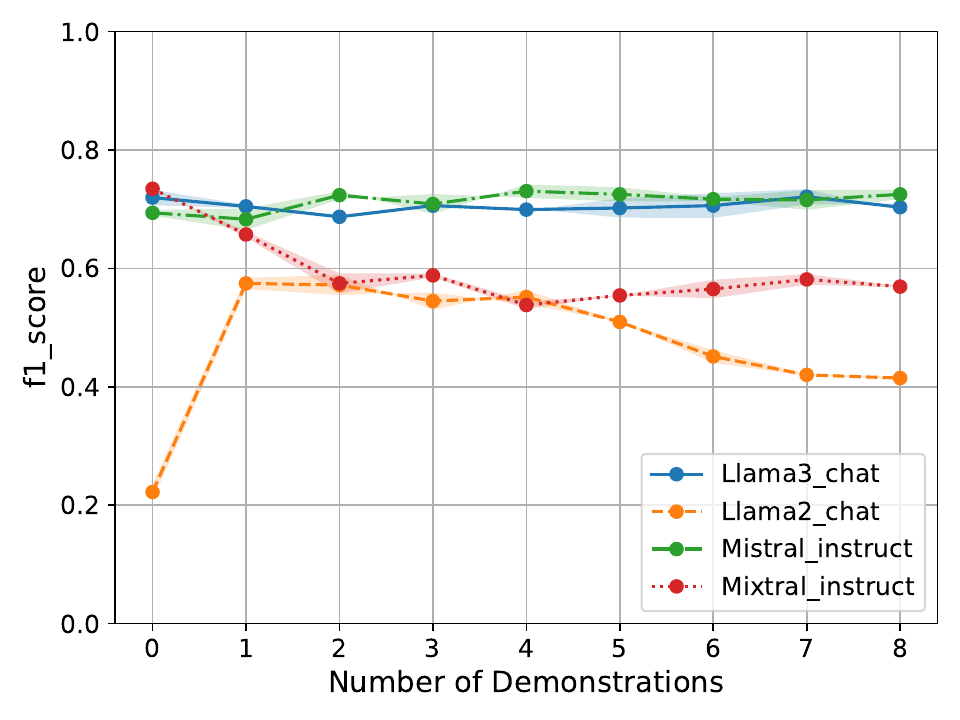}
        \caption*{(e) AMT2 context-aware prompting}
    \end{minipage}
    \hfill
    \begin{minipage}{0.32\textwidth}
        \centering
        \includegraphics[width=\textwidth]{new_figures/pe_sa_f1_context.pdf}
        \caption*{(f) PE context-aware prompting}
    \end{minipage}

    \caption{AR classification. The first row shows the model performance using vanilla prompting on three datasets where the below row shows the performance with the context-aware prompting.}
    \label{fig:ar_classify_app}
\end{figure*}

\begin{figure*}[htb!]
    \centering
    \begin{minipage}{0.32\textwidth}
        \centering
        \includegraphics[width=\textwidth]{new_figures/mt1_adu_f1.pdf}
        \caption*{(a) AMT1 vanilla prompting}
    \end{minipage}
    \hfill
    \begin{minipage}{0.32\textwidth}
        \centering
        \includegraphics[width=\textwidth]{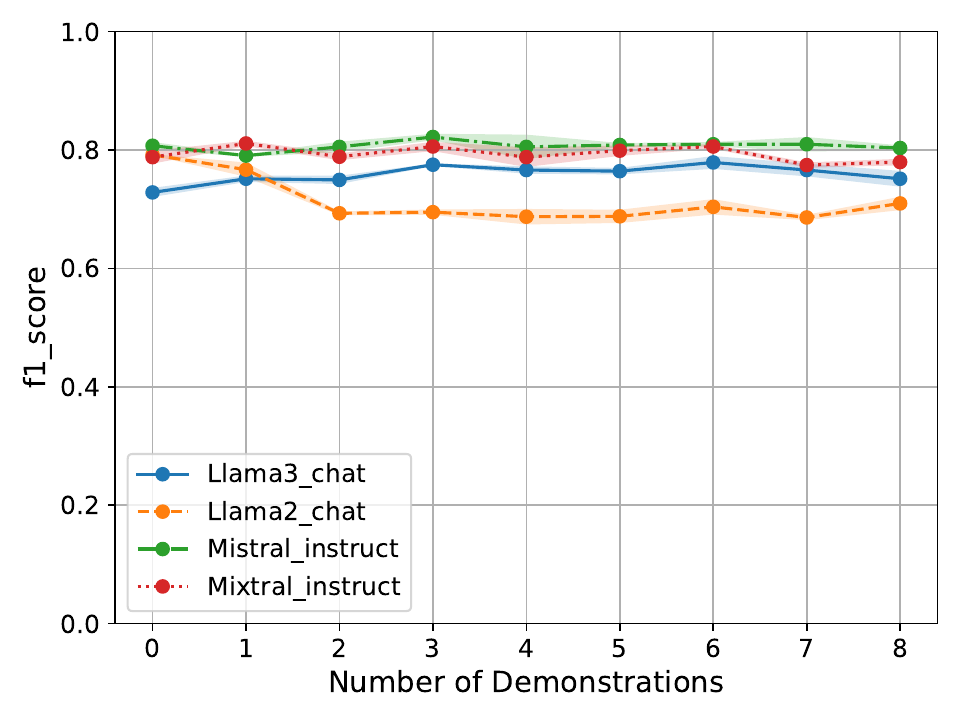}
        \caption*{(b) AMT2 vanilla prompting}
    \end{minipage}
    \hfill
    \begin{minipage}{0.32\textwidth}
        \centering
        \includegraphics[width=\textwidth]{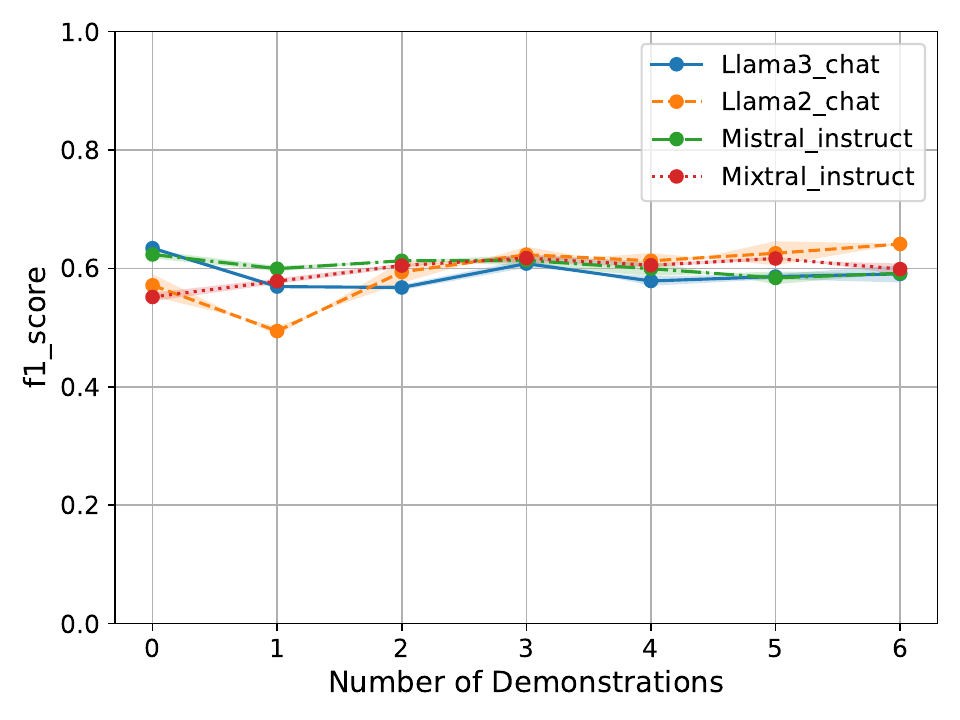}
        \caption*{(c) PE vanilla prompting}
    \end{minipage}
    
    \vspace{10pt} % Adjust space between rows
    \begin{minipage}{0.32\textwidth}
        \centering
        \includegraphics[width=\textwidth]{new_figures/mt1_adu_f1_context.pdf}
        \caption*{(d) AMT1 context-aware prompting}
    \end{minipage}
    \hfill
    \begin{minipage}{0.32\textwidth}
        \centering
        \includegraphics[width=\textwidth]{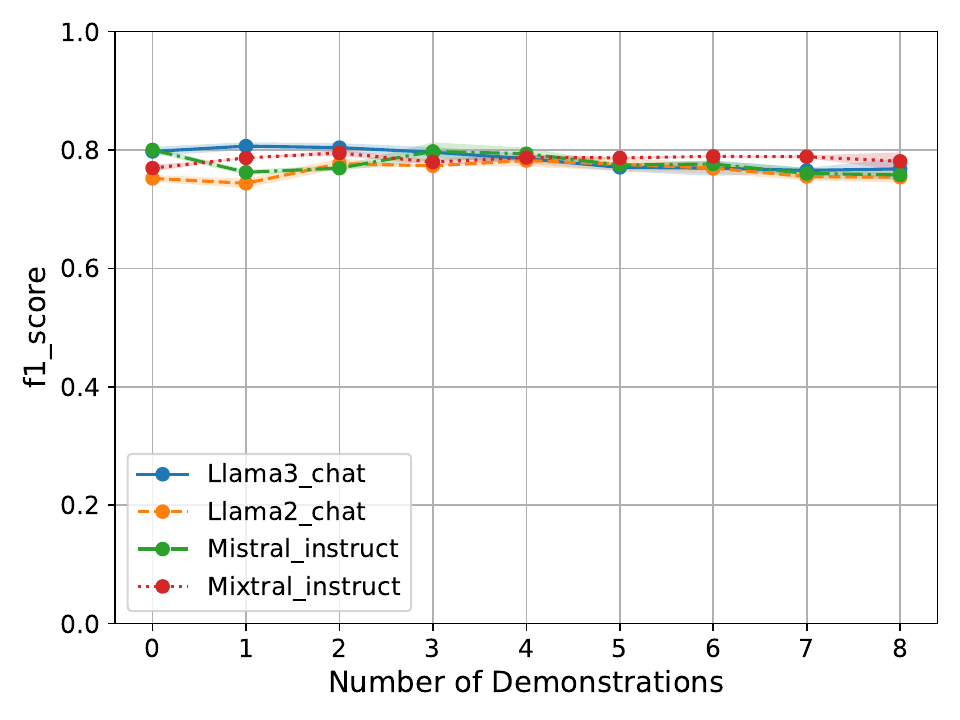}
        \caption*{(e) AMT2 context-aware prompting}
    \end{minipage}
    \hfill
    \begin{minipage}{0.32\textwidth}
        \centering
        \includegraphics[width=\textwidth]{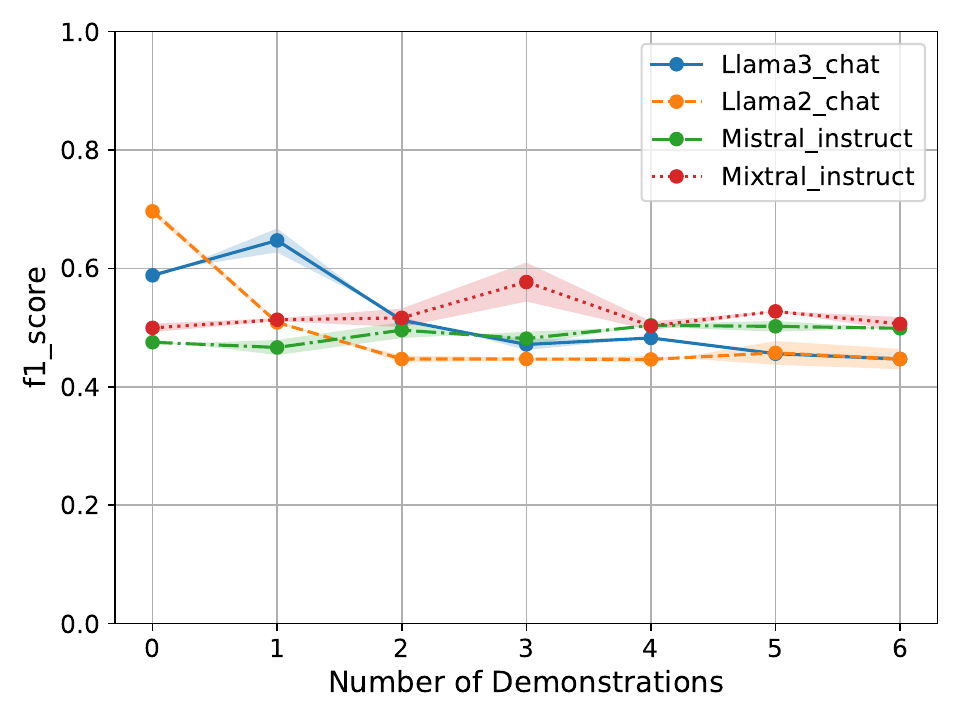}
        \caption*{(f) PE context-aware prompting}
    \end{minipage}

    \caption{ADU classification. The first row shows the model performance using vanilla prompting on three datasets whereas the below row shows the performance with the context-aware prompting.}
    \label{fig:adu_classify_app}
\end{figure*}

\newpage

\section{Prompt Design}
\label{sec:prompt}

In this section, we show four examples of our prompts designed for both vanilla and context-aware prompting methods. 

\begin{table*}[bht!]
  \centering
  \begin{tabular}{p{\linewidth}}
    \hline
    \textbf{System:}  You are an expert in linguistics and you are very good at argumentation mining. Now you are given a paragraph with indexs. Each sub-text is either the claim or premise. Your task is to find the claim in the paragraph. Provide the index of the claim in the text with < >. There is only one correct index.         \\
    \hline
    \textbf{Demo:} Yes, it's annoying and cumbersome to separate your rubbish properly all the time. <2>Three different bin bags stink away in the kitchen and have to be sorted into different wheelie bins. <3>But still Germany produces way too much rubbish, <4>and too many resources are lost when what actually should be separated and recycled is burnt. <5>We Berliners should take the chance and become pioneers in waste separation!          \\
    The answer is: <5>                       \\
    \\
    One can hardly move in Friedrichshain or Neuk\"olln these days without permanently scanning the ground for dog dirt. <2>And when bad luck does strike and you step into one of the many 'land mines' you have to painstakingly scrape the remains off your soles. <3>Higher fines are therefore the right measure against negligent, lazy or simply thoughtless dog owners. <4>Of course, first they'd actually need to be caught in the act by public order officers, <5>but once they have to dig into their pockets, their laziness will sure vanish! \\
    The answer is: <3> \\
    \dots \\
    \hline

    \textbf{Query:} <1>For dog dirt left on the pavement dog owners should by all means pay a bit more. <2>Indeed it's not the fault of the animals, <3>but once you step in it, their excrement seems to stick rather persistently to your soles. \\
    The answer is:
    \\ \hline
  \end{tabular}
  \caption{
    Example of Vanilla Prompting for ADUC task using AMT1 dataset.
  }
  \label{tab:aduc_simple}
\end{table*}

\begin{table*}[bht!]
  \centering
  \begin{tabular}{p{\linewidth}}
    \hline
    \textbf{System:} You are an expert in linguistics and you are very good at argumentation Mining. Now you are given a sentence and a paragraph containing this sentence as a reference. Your task is to classify the sentence as either a Claim or a Premise according to the paragraph. Answer with <0> for Premise and <1> for Claim. There is only one Claim in the paragraph. \\
    \hline
    \textbf{Demo:} Please classify the sentence: Three different bin bags stink away in the kitchen and have to be sorted into different wheelie bins. as either <1> for Claim or <0> for Premise in the given context: Yes, it's annoying and cumbersome to separate your rubbish properly all the time.
    Three different bin bags stink away in the kitchen and have to be sorted into different wheelie bins. But still Germany produces way too much rubbish and too many resources are lost when what actually should be separated and recycled is burnt. We Berliners should take the chance and become pioneers in waste separation! \\
    The answer is: <0>\\
    \\
    Please classify the sentence: And when bad luck does strike and you step into one of the many 'land mines' you have to painstakingly scrape the remains off your soles. as either <1> for Claim or <0> for Premise in the given context: One can hardly move in Friedrichshain or Neuk\u00f6lln these days without permanently scanning the ground for dog dirt. And when bad luck does strike and you step into one of the many 'land mines' you have to painstakingly scrape the remains off your soles. Higher fines are therefore the right measure against negligent, lazy or simply thoughtless dog owners. Of course, first they'd actually need to be caught in the act by public order officers, but once they have to dig into their pockets, their laziness will sure vanish!\\
    The answer is: <0>\\
    \dots \\
    \hline
    \textbf{Query:} Please classify the sentence: For dog dirt left on the pavement dog owners should by all means pay a bit more. as either <1> for Claim or <0> for Premise in the given context: For dog dirt left on the pavement dog owners should by all means pay a bit more. Indeed it's not the fault of the animals, but once you step in it, their excrement seems to stick rather persistently to your soles.\\
    The answer is: 
    \\ \hline
  \end{tabular}
  \caption{
    Example of Context-aware Prompting for ADUC task using AMT1 dataset.
  }
  \label{tab:aduc_context}
\end{table*}

\begin{table*}[bht!]
  \centering
  \begin{tabular}{p{\linewidth}}
    \hline
    \textbf{System:} You are an expert in linguistics and you are very good at Relation Mining.  Now you are given two sentences in an essay. Your task is to classify the relationship between the two sentences as 'Support' if Sentence 1 supports the stance of Sentence 2; or 'Attack' if Sentence 1 does not support Sentence 2. Provide only one word. DO NOT give explanation \\
    \hline
    \textbf{Demo:} Sentence 1:One who is living overseas will of course struggle with loneliness, living away from family and friends. Sentence 2:living and studying overseas is an irreplaceable experience when it comes to learn standing on your own feet. \\
    The answer is: Attack\\
    \\
    Sentence 1:What we acquired from team work is not only how to achieve the same goal with others but more importantly, how to get along with others. Sentence 2:through cooperation, children can learn about interpersonal skills which are significant in the future life of all students.\\
    The answer is: Support\\
    \dots \\
    \hline

    \textbf{Query:} Sentence 1:it also has to be affordable for the consumer. Sentence 2:When a product is commonly used, it becomes trustworthy for the society, no matter what quality it is.\\
    The answer is: \\ \hline
  \end{tabular}
  \caption{
    Example of Vanilla Prompting for ARC task using PE dataset.
  }
  \label{tab:arc_simple}
\end{table*}

\begin{table*}[bht!]
  \centering
  \begin{tabular}{p{\linewidth}}
    \hline
    \textbf{System:} You are an expert in linguistics and you are very good at Relation Mining.  Now you are given two sentences in an essay. Your task is to classify the relationship between the two sentences as 'Support' if Sentence 1 supports the stance of Sentence 2; or 'Attack' if Sentence 1 does not support Sentence 2. Use the context as supporting context. Provide only one word. DO NOT give explanation. \\
    \hline
    \textbf{Demo:}Sentence 1:One who is living overseas will of course struggle with loneliness, living away from family and friends. Sentence 2:living and studying overseas is an irreplaceable experience when it comes to learn standing on your own feet. Please classify the relationship as either Attack or Support based on the given context: Living and studying overseas It is every student's desire to study at a good university and experience a new environment. While some students study and live overseas to achieve this, some prefer to study home because of the difficulties of living and studying overseas. In my opinion, one who studies overseas will gain many skills throughout this experience for several reasons. First, studying at an overseas university gives individuals the opportunity to improve social skills by interacting and communicating with students from different origins and cultures. Compared to \dots \dots in general life. \\
    The answer is: Attack\\
    \\
    Sentence 1:What we acquired from team work is not only how to achieve the same goal with others but more importantly, how to get along with others. Sentence 2:through cooperation, children can learn about interpersonal skills which are significant in the future life of all students. Please classify the relationship as either Attack or Support based on the given context: Should students be taught to compete or to cooperate?It is always said that competition can effectively promote the development of economy. In order to survive in the competition, companies continue to improve their products and service, and as a result, the whole society prospers. However, when we discuss the issue of competition or cooperation, \dots \dots in one's success. \\
    The answer is: Support \\
    \dots \\
    \hline

    \textbf{Query:} Sentence 1:it is necessary to make sure that people can live a long life. Sentence 2:animal experiments have negative impact on the natural balance. Please classify the relationship as either Attack or Support based on the given context: Using animals for the benefit of the human beings with the rapid development of the standard of people's life, increasing numbers of animal experiments are done, new medicines and foods, for instance. Some opponents says that it is cruel to animals and nature, however, I believe that no sensible person will deny that it is a dramatically cruel activity to humanity if the latest foods or medicines are allowed to be sold without testing on animals. In my essay, I will discuss this issue from twofold aspects.First of all, as we all know, animals are friendly and vital for people, because if there are no animals in the world, the balance of nature will break down, and we, human, will die out as well. The animal experiments accelerate the vanishing of some categories of animals. In other words, doing this various testing is a hazard of human's future and next generation.Though animal experiments have negative impact on the natural balance, it is necessary to make sure that people can live a long life. To begin with, it is indisputable that every new kind food or pill may be noxious, and scientists must do something to insure that the new invention benefits people instead of making people ill or even dying. The new foods or medicines are invented to promote the quantity of human's life. Thus even if they are volunteers; they cannot take the place of animals to test the new foods or medicines. Furthermore, it also have potentially harm for human's health without any testing. To sum up, I reaffirm that although there is some disadvantages of animals' profits, the merits of animal experiments still outweigh the demerits. \\
    The answer is: \\ \hline
  \end{tabular}
  \caption{
    Example of Context-aware Prompting for ARC task using PE dataset.
  }
  \label{tab:arc_context}
\end{table*}

\end{document}